\documentclass[journal]{IEEEtai}

\usepackage{graphicx}

\usepackage[american]{babel}

\usepackage{booktabs}

\usepackage{subcaption}
\usepackage{mathtools}
\usepackage{svg}
\usepackage{float}
\usepackage{balance}

\usepackage{amsfonts}

\usepackage{algorithm}
\usepackage{algcompatible}

\newcommand{\MethodnameLong}{Federated Learning for distribution skewed data using sample weights}
\newcommand{\MethodnameShort}{FedDisk}

\newcommand{\px}[2]{p_{#1}(\mathbf{x_{#2}})}
\newcommand{\qx}[2]{q_{#1}(\mathbf{x_{#2}})}

\newcommand{\Prob}{\mathcal{P} }
\newcommand{\prob}{\mathcal{P} }

\newcommand{\x}{\mathbf{x}}
\newcommand{\ubold}{\mathbf{u}}

\newcommand\parties{100}

\usepackage{stmaryrd}

\usepackage{multicol}

\usepackage[space]{grffile}

\usepackage{clipboard}
\newclipboard{myclipboard}
\usepackage{xcolor, soul} 
\definecolor{OliveGreen}{cmyk}{0.64,0,0.95,0.40}
\definecolor{marygold}{cmyk}{0,0.1,0.5,0}

\begin{document}

\title{\MethodnameLong}

\author{Hung~Nguyen~\IEEEmembership{Member,~IEEE,}
    Morris~Chang~\IEEEmembership{Member,~IEEE,}
    Peiyuan~Wu~\IEEEmembership{Member,~IEEE,}
}

\author{\IEEEauthorblockN{Hung Nguyen\IEEEauthorrefmark{1}
		,~Peiyuan~Wu~\IEEEauthorrefmark{2} 
		and ~J. Morris Chang\IEEEauthorrefmark{3}}
		
	\IEEEauthorblockA{\IEEEauthorrefmark{1}\IEEEauthorrefmark{3} Electrical Engineering Dept, University of South Florida, USA\\
	\IEEEauthorrefmark{2} Electrical Engineering Dept, National Taiwan University, Taiwan \\
		Email: \IEEEauthorrefmark{1}nsh@usf.edu}
	
	\thanks{Code and Data are available at https://github.com/nsh135/FedDiskPytorch}
	}

\maketitle
	
\begin{abstract}
\Copy{non-IID}{
   One of the most challenging issues in federated learning is that the data is often not independent and identically distributed (non-IID). Clients are expected to contribute the same type of data and drawn from one global distribution. However, data are often collected in different ways from different resources. Thus, the data distributions among clients might be different from the underlying global distribution. This creates a weight divergence issue and reduces federated learning performance. This work focuses on improving federated learning performance for skewed data distribution across clients.} The main idea is to adjust the client distribution closer to the global distribution using sample weights. Thus, the machine learning model converges faster with higher accuracy. We start from the fundamental concept of empirical risk minimization and theoretically derive a solution for adjusting the distribution skewness using sample weights. To determine sample weights, we implicitly exchange density information by leveraging a neural network-based density estimation model, MADE. The clients' data distribution can then be adjusted without exposing their raw data. Our experiment results on three real-world datasets show that the proposed method not only improves federated learning accuracy but also significantly reduces communication costs compared to the other experimental methods.
   
\end{abstract}

\begin{IEEEImpStatement}
	Non-IID issue is a well-known problem in machine learning applications, especially for Federated Learning, as clients often collect data from different sources and in different conditions. The problem significantly reduces machine learning performance and increases communication costs. To alleviate the nagative impact of non-IID data, several works have been proposed. However, they mostly require clients to share part of their private data or rely on the global information from the global model, which already suffered the non-IID. In this work, the proposed method adjusts the distribution via sample weight in the loss function during training. We only ask clients to share some extra models. Similar to a typical FL framework, clients are not required to expose their raw data. The results on three real-world datasets showed that the proposed method is much more efficient than other experimental methods. It increases the accuracy of ML model and significantly reduces communication costs e.i., up to eight times for real non-IID dataset FEMNIST.     
\end{IEEEImpStatement}	
	
\begin{IEEEkeywords}
feature skewness, communication cost reduction, privacy preservation, deep learning.
\end{IEEEkeywords}

\IEEEpeerreviewmaketitle

\section{Introduction}

Since the demand for massive data in artificial intelligent machines, federated learning (FL) was first introduced in 2016, \cite{OriginFL}, a collaboratively decentralized learning framework, in contrast to centralized learning approaches (in which datasets are sent to an aggregator), FL encourages data holders to contribute without the privacy concern of exposing their raw data. For example, a number of hospitals holding patient records would participate in a machine learning (ML) system to provide better disease predictions via a FL framework without the concern of privacy disclosure. An FL framework is then set up to train a global machine learning model with the participation of all hospitals (clients) and an aggregator coordinating the model transfer between clients and the aggregator. Instead of sharing raw data, each client trains their ML model for a small number of iterations and updates the model parameters to the aggregator. The aggregator then merges all clients' model parameters into a new global model and sends them back to the clients for further training iterations. Clients receive the global model and continue to train for the next iterations. The process is repeated until the global model is fully trained.    

Because the data in such collaborative learning usually come from different sources, they might be drawn from local distributions that are different from the underlying global distribution. This can be considered a non-IID issue, and it might cause a performance reduction. \Copy{para:motivation}{The primary problem is the divergence of model weights, as found in \cite{Zhao2018FederatedLW} by Zhao et al. The authors showed that the model's weights tend to be more diverged for non-IID data compared to that for IDD data. This causes a performance reduction, and it worsens as the data distribution becomes more skewed. For example, the accuracy dropped by about 10\% for image dataset Cifar-10 \cite{cifar10}, and speech recognition dataset KWS \cite{kws} with a non-IID setting}. In FL, this issue arises due to differences between client individual distribution and the global distribution. \Copy{sec:introduction}{To address this problem, we proposed an algorithm that utilizes sample weights to adjust individual client distributions closer to the global distribution during the training process. However, obtaining global information across clients is challenging in an FL setting because clients need to allow the exposure of their raw data. To overcome this challenge, the proposed method implicitly shares statistical information of client data without revealing the client's raw data. The method only requires clients to exchange additional model weights using a typical FL procedure. Once the adjustment weights are acquired, the machine learning model can be trained using a standard FL framework. The proposed method is demonstrated to improve FL accuracy and significantly reduce FL communication costs through experiments on three real-world datasets.

Our contributions are as follows:
\begin{enumerate}
\item Provide a theoretical base for skewed feature distribution data for federated learning by adjusting sample weights derived from the machine learning empirical risk.  
\item Provide a practical solution to mitigate the problem of learning from non-IID data for the FL framework without sharing clients' draw data. The proposed method only requires clients to share additional model parameters, similar to a typical federated learning framework.
\item Several experiments were conducted on three datasets, including MNIST, non-IID benchmark dataset FEMNIST and real-world dataset Chest-Xray. The results demonstrate that the proposed method outperforms other experimental methods in classification accuracy and dramatically reduces the communication cost.
\item As the proposed method needs to exchange additional information, we also provide a theoretical analysis to analyze the potential privacy leakage. We showed that the leakage information becomes insignificant when the number of clients increases. 
\item To our best knowledge, the proposed method is the first method utilizing data distribution information and sample weights to tackle the FL Non-IID issue.   
\end{enumerate}
}

The rest of this paper is organized as follows. A brief review was conducted in Section \ref{sec:relatedwork} Related Work. Section \ref{sec:problem} formulates our problem and our goal to achieve. Section \ref{sec:made} introduces a neural network-based model that is leveraged in our work to carry density information. Our proposed solution is introduced in Section \ref{sec:methodology}. A comprehensive study on the modules in the proposed method is introduced in Section \ref{sec:ablation} Ablation Study. We provide a privacy leakage analysis in Section \ref{sec:privacyAnalysis} as the proposed method indirectly exchanges distribution information. Section \ref{sec:experiments} shows our experimental results and illustrates the proposed method's performance. Section \ref{sec:conclusion} summarizes our study and discusses future work to improve the proposed method.     

\section{Related Works}
\label{sec:relatedwork}
To learn a model utilizing data from multiple clients without directly accessing clients' data, authors in \cite{OriginFL} first introduced Federated Averaging (FedAvg) and demonstrated its robustness in 2016. The main idea is that clients (data holders) are involved in a model training process by exchanging local models' parameters instead of exchanging raw data. Since then, FL has been seen in various applications in different fields \cite{abs-1811-03604,yadav_federated_2022,feki_federated_2021,zhai_dynamic_2021, distributedQuantum, hierarchicalFL}. Federated Learning can be categorized into two primary scenarios: cross-silo and cross-device FL. In the former, there are fewer clients, each having substantial data (such as hospitals). In contrast, the latter scenario involves numerous clients with lightweight devices, and each client may have a smaller dataset (like mobile users). One of the main concerns in FL is that the data might come from different sources and have different distributions. Thus, FL performance is significantly reduced because this violates a fundamental machine learning assumption that data should be independent and identically distributed (IID). The FL over non-IID data has been shown in existing works \cite{ZHU2021371,Sahu2018OnTC,9392310,abs_1905_06641,Shen2020FederatedML,9155494,abs-2102-02079,abs-2005-11418,FLviaIntel} that its performance deteriorates dramatically. In this work, we focus on tackling the non-IID data issue in a federated learning system in which the collected data feature distribution is skewed. Many different reasons might cause the skewness. For example, clients might perform different sampling methods, apply different normalization methods, or sample using different devices. 

Over the past few years, there have been a number of approaches aiming at reducing non-IID data impacts. While many current works focus on skewed label distribution, there are only limited approaches considering skewed feature distribution data, which is very common in various fields, e.g., medical images collected from different X-ray machines. We classify the existing works into three categories as follows.

\begin{itemize}
\item \textbf{Sharing data:}
This approach mainly focuses on adjusting model weights or calibrating model parameters using sharing data. However, they require a certain amount of raw data to be shared among users. For example, authors in \cite{Zhao2018FederatedLW, nofearofheterogeneity} proposed an alleviation by finetuning the model on globally shared data to adjust the distribution drift. Thus, this still poses a privacy concern. Zhu et al. in \cite{distillationFL} used adversarial training and shared the generator for generating synthetic samples, which might contain global information. However, training generators also suffered from the non-IID itself.       

\item \textbf{Training stabilization:}
This approach focuses on stabilizing the local training process by regulating the deviation between local parameters and global parameters. This could be implemented by normalizing layers, adding regularization terms in loss functions, or sharing specific layers. For example, Li et al. illustrated in their work (FedBN) \cite{li2021fedbn} that Local Batch Normalization would help to reduce the problem of non-IID data. FedBN suggests clients not synchronize local batch normalization parameters with the global model. Sahu et al. introduce FedProx \cite{FedProx} to solve the weight-divergence issue by proposing a loss function that constrains the local models to stay close to the global model. FedNova \cite{fednova} suggests to normalize local weights before synchronizing with the aggregator. FedMA \cite{fedma}, AFL \cite{AFL}, and PFNM \cite{pfnm} consider combinations of layer-wise parameters and provide an aggregation of such parameters to alleviate the non-IID issue. In FedRod \cite{FedRod}, Chen and Chao deal with the non-IID issue by learning hyper-networks locally, which results in personalized classifiers for clients and clients' class distributions. Recently, Tan et al. \cite{fedpcl} tackled the non-IID data issue by exchanging representation vectors of samples in a given class instead of the model's parameters, enabling clients to have personalized model architecture.    
 
\item \textbf{Weighted aggregation:}
The approach mitigates heterogeneity by adjusting the model weight during aggregation. However, this works on the model level, so it is ineffective if the data is non-IID within a client. FedCL \cite{fedcl} and FedDNA \cite{fedDNA} share statistical parameters of models (means and variances) and aim at finding averaging weights for each client's model to minimize models' weights divergence across clients. However, as this approach only considers the aggregating weights for each model, the improvement is minor. 
\end{itemize}

\Copy{par:problemStatement}{
\section{Problem Statement}
\label{sec:problem}
In this section, we introduce and formulate the scenario of FL with skewed feature distribution across clients. 
Our scenario is a learning collaboration between $K$ clients to build a global classification model that maximizes the global accuracy given arbitrary data. Each client holds a number of individual records that they are not willing to share with others due to privacy concerns. This study focuses on preventing the performance of the global model from deteriorating because of the distribution skewness issue \cite{survey} across clients.

Our goal is to adjust the clients' distributions to be closer to the global distribution via sample weights. We denote the data and associated labels held by client $k$ $\in$ $\{1,...,K\}$ as $ \{( \mathbf{x}_k^j,y_k^j )\}_{j=1}^{N_k}$ where $ \mathbf{x}_k^j \in \mathbb{R}^d$ and $y_k^j \in \mathbb{N}$. Let the data distribution of the $k^{th}$ client be $q_k(\mathbf{x})$ and the ground truth global distribution is $p(\mathbf{x})$. Our problem becomes finding an adjusting weight function for each client $k$, $\alpha(\mathbf{x}_k)$ such that
	
\begin{equation}
	\alpha_k(\mathbf{x}) q_k(\mathbf{x}) = p(\mathbf{x})
	\label{equ:goal}
\end{equation} 
}

\section{Preliminary: Masked Autoencoder for Distribution Estimation (MADE) }
\label{sec:made}
The proposed method asks the clients to share additional model weights that carry their local datasets' distribution information instead of sharing the raw data. We utilize a neural network-based density estimation, namely, Masked Autoencoder for Distribution Estimation (MADE) \cite{MADE}. This section briefly introduces MADE.

MADE is designed to estimate the probability distribution of input components (e.g., pixels in an image). MADE assumes input components are dependent instead of independent, which is relevant in many applications. For example, MADE can decompose the distribution of an instance $\textbf{x}$ consisting $n$ components $x_1, x_2, x_3, ..., x_n$ as follows:
\begin{equation}
	\label{eq:MADE_decomposition}
   p(\mathbf{x}) =   p(x_1|x_2,x_3,..,x_n) \cdot p(x_2|x_3,...,x_n) 
					...p(x_{n-1}|x_n) \cdot p(x_n). 
\end{equation} In our study, the instances are images and each pixel can be considered as a component. Thus, $n$ is the size of a flatten image vector. 
 
For MADE implementation, a shallow neural network is utilized. Its input and output size are equal (similar to an Autoencoder), for example a size of $n$ for the above example. The main idea is to mimic Equation \ref{eq:MADE_decomposition} by masking neuron connections across layers to control the seen and unseen connections to model output. Specifically, MADE poses constraints on the model that each output component in a certain layer only connects to its dependent input components in the previous layer. Masks are created based on such principle, and applied to the weights of the model.

Specifically, MADE assigns each unit in a hidden layer an integer $\textit{m}$ between $1$ and $D-1$, where $D$ is the number of dimensions. Denote $\textit{m(k) } $ as the maximum number of units in the previous layer to which the $\textit{k}^{th}$ hidden unit can connect, the weight mask $M$ is then formulated as follows:
$
 M_{k,d} = 1_{m(k) \geq d} = 
				 \begin{cases}
				  1 & \text{if \textit{m(k)} $\geq$ d }\\
				  0 & \text{otherwise,}
			 \end{cases} 
$
for $d \in \{1,...,D\}$ and $k \in \{1,...,K \}$ with $K$ being the number of hidden layer units.

\section{\MethodnameLong{}}
\label{sec:methodology}
\Copy{par:howEffectiveness}{
In this section, a solution is proposed to alleviate the negative impact of distribution skewness across clients for federated learning by adjusting client data distribution during the training process. The proposed method aims to find weights for training samples in order to adjust client data distributions. The remainder of this section introduces how we design sample weights. We also show that the goal in Equation \ref{equ:goal} can be derived from the machine learning optimization problem as described in this section. 

By applying the sample weights for the local training on each client, the proposed method reduces the distribution skewness of each client's data and prevents clients' raw data from being exposed. Some statistical information between clients and the aggregator must be exchanged to find sample weights. However, instead of exchanging the raw information, which might hurt clients' privacy, the proposed method only exchanges model parameters, similar to a typical Fl framework.}

\begin{figure*}[ht!]
		\centering
		\includegraphics[width=0.85\textwidth, trim={1cm 0.1cm 1cm 0.3cm},clip]{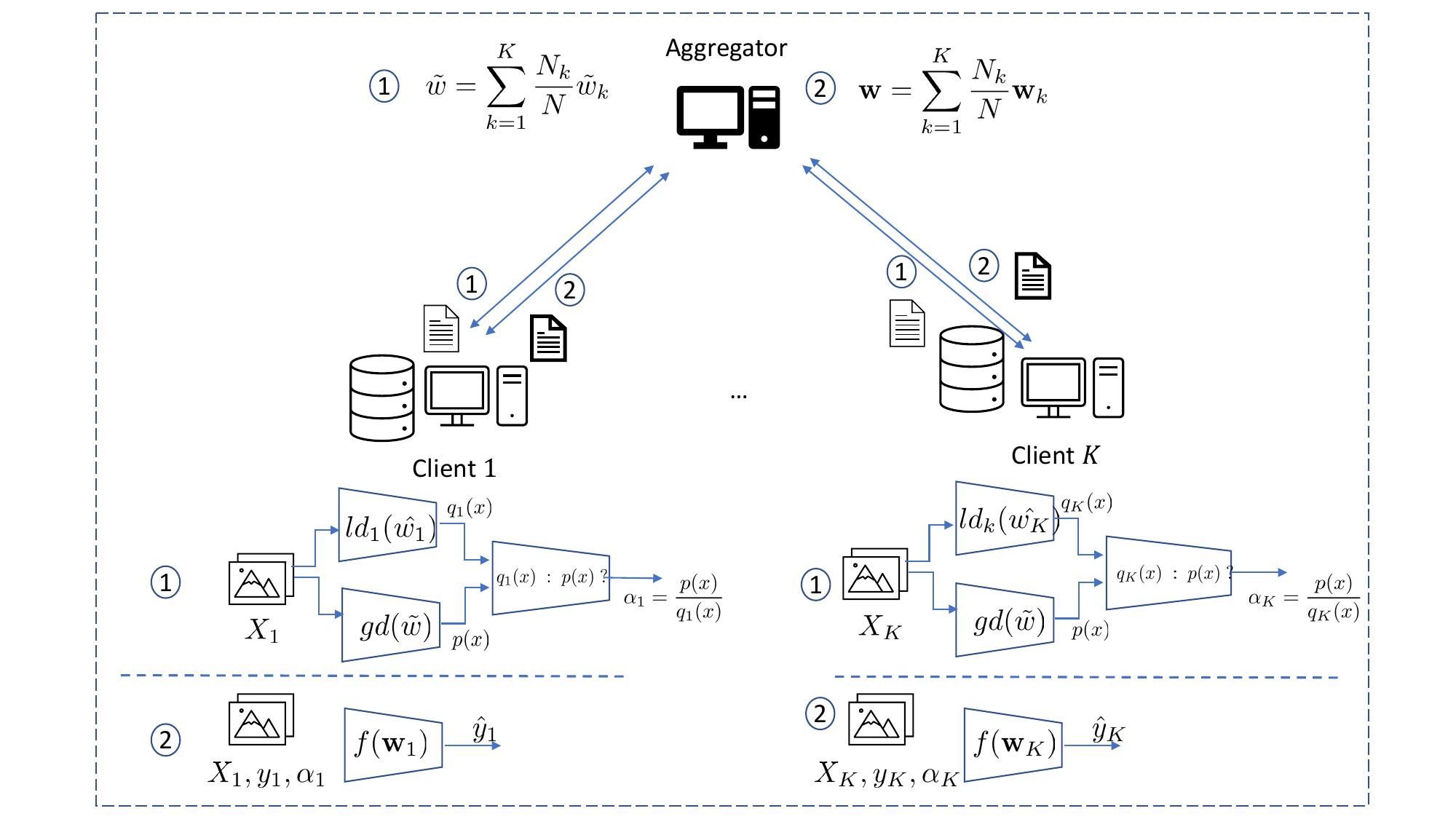}
		\caption{ \MethodnameShort{} Framework: The proposed framework has two phases. First, local and global probability density functions ($p(x),q(x)$) are estimated via MADE models leveraging FL procedures. Then, the sample weights $\alpha$ are computed by approximating density ratio via class probability estimation. Second, the machine learning tasks (e.g., classification) can be performed similar to a typical FL method (i.e., FedAvg) with the sample weights acquired from phase 1.} 
		\label{fig:framework} 

\end{figure*}

Our framework is illustrated in Figure \ref{fig:framework}. Where $f(\cdot)$ denotes local inference models and $\mathbf{w}_k$ is the model's parameter of the $k^{th}$ client. The proposed method, namely \MethodnameShort{}, requires a 2-phase process. First, clients jointly learn a global density estimation model and their local density models utilizing MADE models. These models are then used to derive sample weights for the local training process. Second, the machine learning tasks can be learned by the conventional FL procedure, with the data skewness issue mitigated by the sample weights from the first phase.

\subsection{Sample Weights Design}
\label{sec:sample_weight_design}
As we do not have sufficient information about the true distribution, we consider the combination of all clients' dataset distribution as our true distribution. Thus, we consider the probability density function (pdf) of the true distribution as 
\begin{equation}
\label{eq:sumdist}
	 p(\mathbf{x}) = \sum_{k=1}^{K} c_k q_k(\mathbf{x}) ,
\end{equation}
where $q_k(\mathbf{x})$ denotes the pdf of the $k^{th}$ client's data. $c_k$ depicts the client's weight, as determined by the ratio of the client's sample count $N_k$ to the total number of samples $N$: 
\begin{equation}
	c_k=\frac{N_k}{N}.
\end{equation}

To jointly learn a global model, the system finds the expectation of the loss function $l(g(\mathbf{x}),y)$ with sample $\x{}$ that drawn from the true distribution. The expected loss is formulated by the associated risk \cite{noauthor_empirical_2021} as follows:
\begin{align}
	\mathbb{E}[l(g(\mathbf{x}),y)] 
	&= \iint l(g(\mathbf{x}),y) p(y|\mathbf{\mathbf{x}}) p(\mathbf{x}) d\mathbf{x}dy, \label{eq:risk} 
\end{align}
where $p(\mathbf{x},y)$ is the joint pdf of a sample $\mathbf{x}$ and its associated label $y$, and $p(y|\mathbf{x})$ is the conditional probability of a label $y$ given a sample $\mathbf{x}$. We also assume that for any client $k \in$ $\{1,...,K\}$ with local data distribution $q_k(\mathbf{x})$, the conditional probability of a label $y$ given a sample $\mathbf{x}$ is equivalent to that of the true distribution, namely
\begin{align}
	\label{eq:assumtion}
	 q_k(y|\mathbf{x}) = p(y|\mathbf{x}).
\end{align}

From Equation \ref{eq:sumdist}, \ref{eq:risk}, \ref{eq:assumtion}, and by multiplying with factor $\frac{q_k(\mathbf{x})}{q_k(\mathbf{x})}=1$, the expected loss in Equation \ref{eq:risk} can be expanded as follows:
\begin{align}
	\mathbb{E}[l(g(\mathbf{x}),y)]  &= \iint l(g(\mathbf{x}),y) p(y|\mathbf{x}) p(x) d\mathbf{x}dy,\\
	&= \iint l(g(\mathbf{x}),y) q_k(y|\mathbf{x}) \frac{q_k(\mathbf{x})}{q_k(\mathbf{x})}  p(\mathbf{x}) d\mathbf{x}dy \\
	&= \iint l(g(\mathbf{x}),y) q_k(\mathbf{x},y) \frac{p(\mathbf{x})}{q_k(\mathbf{x})}  d\mathbf{x}dy.    \label{eq:expectedloss}
\end{align}

The objective of the global model thus amounts to minimize the empirical risk over all $K$ clients' datasets:
\begin{equation}
	\begin{aligned}
	\underset{g}{\text{minimize}}
	 \frac{1}{N}\sum_{k=1}^{K} \sum_{j=1}^{N_k} \alpha_k^j \:l(g(\mathbf{x}_k^j),y_k^j)), \label{eq:optimation}
	\end{aligned}
\end{equation} 

where $\mathbf{x}_k^j$, $y_k^j$ are the $j^{th}$ sample and its label. $N$ is the total number of samples. $\alpha_k^j$ denotes $\alpha_k(\mathbf{x}^j)$, which represents client $k$ sample weight function with respect to $\mathbf{x}$, computed as
\begin{equation}
\begin{aligned}
	\boxed{\alpha_k^j = \frac{ \px{}{}}{q_k(\mathbf{x})}  = \frac{ \sum_{i=1}^{K}q_i(\mathbf{x}) }{q_k(\mathbf{x})}. }  
	\label{eq:alpha}.
\end{aligned}
\end{equation}

Our problem becomes minimizing the summation of the loss functions (Equation \ref{eq:optimation}) over all clients. For each client, the loss function is minimized over local samples with the corresponding $j^{th}$ sample weight of the client $k^{th}$, $\alpha_k^j$. The sample weights could be estimated by the density ratio between the true distribution (global distribution) and the client distributions (local distributions). For each client, the local distribution can be estimated using its local data. However, the challenge is to achieve the true distribution without having access to other clients' data. To solve this, we leverage a neural network-based density estimation model to learn the global density function via a typical federated learning procedure. Thus, clients can implicitly exchange some statistical information, while still preserving the privacy in client data.          

\subsection{Probability Density Approximation}
\label{sec:densityapproximation}
To estimate global density and preserve client privacy at the same time, we propose to leverage a neural network-based density estimation so that we can exchange local density information (via models' weights) with the aggregator without sharing the raw data. In this work, we leverage a well-known method, namely, Masked Autoencoder for Distribution Estimation (MADE, \cite{MADE}), which is briefly reviewed in Section \ref{sec:made}. Elaborately, each client aims to estimate its local probability density $q_k(\mathbf{x})$ using its own dataset, and all $K$ clients jointly estimate the global probability density $p(\mathbf{x}) = q_1(\mathbf{x})+...+q_K(\mathbf{x})$. Learned MADE models are used to approximate local probability density functions, and the global MADE model approximates the global probability density. The learning process is described as follows.    

The $k^{th}$ client learns a local density estimation model $ld_k(\hat{w}_k)$ (where $ld(\cdot)$ approximates density estimation function with parameter $\hat{w}$) using its local data. It then jointly learns a global density estimation model $gd(\tilde{w})$ (where $gd(\cdot)$ represents the global density function with the parameter $\tilde{w}$) using the procedure similarly to FedAvg \cite{OriginFL}. Specifically, for the local model density estimation models, each client train a MADE model on its local data until the loss function can not be improved. For the global density estimation model, each client trains its data locally for a certain number of iterations, and then model parameters are sent to an aggregator for the aggregation. Since clients might own different number of samples, a weight of ${N_k}/N$ (where $N_k$ and $N$ are the number of samples of the $k^{th}$ client and the the total number of samples over all clients) is used for adjusting client parameter significance, similar to FedAvg. After aggregating all clients' model parameters, the aggregator shares global model parameters to all clients. The steps are repeated until the validation loss starts increasing. The global MADE model aggregation from $K$ clients at iteration $t$ can be described as follows:
\begin{equation}
\begin{aligned}
 \tilde{w}^t = \sum_{k=1}^{K} \frac{N_k}{N} \tilde{w}^t_k 
\end{aligned}
\end{equation}

\subsection{Sample Weight Approximation}
\label{sec:weightapproximation}
After the local and global density approximations by MADE models are fully learned ( Section \ref{sec:densityapproximation}), we can estimate sample weights in Equation \ref{eq:alpha}. Since MADE models output vectors of conditional probabilities for each element in the d-dimensional input $\mathbf{x}$, an intuitive way to compute $p(\mathbf{x})$ is to multiply all the conditional probabilities. However, as $\px{}{}$ vanishes when any of the conditional probabilities vanishes, we instead keep the output as a vector of conditional probabilities (same size as input) and approximate the density ratio in Equation \ref{eq:alpha} using a class probability estimation method inspired by \cite{densityratio}. The method aims at training a binary classifier to output a probability that represents the ratio between $\px{}{}$ and $\qx{}{}$. The solution detail is described in the rest of this subsection. 

After each client receives the final global MADE model and trains its own local MADE, it starts to evaluate sample weights for its local data. The training data of the $k^{th}$ client, $X_k$, is then fed into both the global MADE (the global MADE is downloaded to clients so that this step can be done locally) and the local MADE to estimate $\px{}{}$ and $\qx{k}{}$, respectively. Denote $\ubold{}$ as the output vector of density estimation models, and $l$ be the pseudo label indicating whether it is sampled from the global destination ($l=1$) or the local distribution ($l=0$). Each client then trains a binary classifier to differentiate whether the output $\ubold$ comes from $\px{}{}$ or $\qx{k}{}$. Outputs of the two MADE models (the sample size of each output is $N_k$) are concatenated to a new vector dataset including samples $\{(\ubold{}_k^i,l_k^i) \}_{i=1}^{2N_k}$, and is used to train the binary classifier. The conditional probabilities of the binary classification model $h(\ubold,w_h)$ (where $\ubold$ is the input variable , $w_h$ is the model parameter) can be approximated as following: 
\begin{align}
	\Prob(\ubold | l=0 )  \propto  \qx{k}{}  ,\;\;\;\;
    \Prob(\ubold | l=1 )  \propto \px{}{}.
\end{align}

From Bayes' rule, we have
\small
\begin{align}
	\frac{\px{}{}}{\qx{}{}} &= \frac{\Prob(\ubold | l=1 )}{ \Prob(\ubold | l=0 ) }       
	= \left( \frac{\Prob(l=1| \ubold ) \Prob(\ubold)}{\Prob(l=1)} \right)     
	\left( \frac{\Prob(l=0)} {\Prob(l=0| \ubold ) \Prob(\ubold)}\right)  \label{eq:pq1}\\
	 &= \frac{  \Prob(l=1|\ubold )  \Prob(l=0) }{ \Prob(l=0|\ubold )  \Prob(l=1)  }.  \label{eq:pq2} 
\end{align}

We approximate the marginal probability ratio between two distributions ( $\Prob(l=0)$  and $\Prob(l=1)$) by the number of samples from the two distributions $N_k$ over the concatenated dataset size ($2N_k$).Thus,  We have
 	$\frac{ \Prob(l=0) }{ \Prob(l=1)  } = \frac{N_k}{2N_k} \frac{2N_k}{N_k}=1. \label{eq:marginalratio}$

The density ratio then can be estimated as follows:
\begin{align}
\frac{\px{}{}}{\qx{}{}} &= \frac{  \Prob(l=1|\ubold )  }{ \Prob(l=0|\ubold )   } 
	= \frac{  \Prob(l=1|\ubold )  }{ 1 - \Prob(l=1|\ubold )   } ,\label{eq:ratio2}
\end{align} 
where $\Prob(l=1|\ubold{})$ is the classifier's probability-liked output indicating how likely an input vector $\ubold$ comes from the global probability $\px{}{}$. 
 
To summarize, the $j^{th}$ training sample of client $k$, $\x_k^j$, is fed into the client's local MADE model to achieve its corresponding density estimation $\ubold{}_k^j$. $\ubold{}_k^j$ is then fed into the binary classification function $h(\ubold)$ to achieve the class probability $ \Prob(l=1|\ubold_k^j ) $. This is used to estimate the sample weight $\alpha_k^j$ (Equation \ref{eq:alpha}) based on Equation \ref{eq:ratio2}. In words, the binary classification model $h(\ubold,\bar{w})$ is expected to return higher weights for samples that are likely belonging to the true distribution and vise versa. 

\subsection{Learning With Skewed Distribution Data Across clients }
After acquiring sample weights, each client starts to train the model on the local dataset and corresponding sample weights for a machine learning task (e.g., classification) as a typical FL framework. In this work, we follow the procedure introduced by FedAvg to learn the global model. The aggregator aggregates clients' local models as follows:
\begin{equation}
\begin{aligned}
\mathbf{w}^t = \sum_{k=1}^{K} \frac{N_k}{N} \mathbf{w}^t_k 
\end{aligned}
\end{equation}
where $\mathbf{w}^t$ and $\mathbf{w}^t_k$ are the global and local model parameter of $k^{th}$ client at the $t^{th}$ iteration.

\Copy{sec:algorithm}{
\subsection{Algorithm}
\label{sec:algorithm}
Our main strategy can be described in Algorithm \ref{alg:FedDisk}. We only concentrated on describing the first phase of FedDisk as the second is the same as a typical FL framework. First, each client trains its own local MADE model on the local data to obtain local distribution information (lines $1-3$). All clients then jointly train global MADE utilizing a typical FL framework (lines $4-14$). After achieving these two models, data are sampled from the two output models to acquire data samples containing local and global information (lines $15-17$). These samples are concatenated with the pseudo label of $0$ for samples that come from local distribution and $1$ for ones from the global distribution (line $18$). They are then used for training an adversarial binary classifier (line $19$). The purpose is to differentiate the two datasets sampled from the two distributions. The samples that are similar to the global samples will return a higher probability of belonging to class $1$ (come from global distribution) and vice versa. Thus, the classifier probability-like output that represents class $1$ is then used to be the weights for the sample (line $20$).  

\begin{algorithm}[ht!]
	\caption{Phase 1: Sample Weight Computing.}
	\begin{flushleft}
		\textbf{Input}: 
		Client $k^{th}$: Dataset \{$X_k,y_k$\}.\\
		\textbf{Parameter}: 
		$K$ : Number of client.\\
		$N$ : Number of total samples.\\
		$N_k$: Number of sample of client $k^{th}$. $\sum_{k=1}^{K}N_k = N$\\
		$local_iter$: Number of iterations for training local model\\
		$global_iter$ : Number of iterations for training global model.\\
		$ld(\hat{w})$ : Local MADE model containing \textbf{l}ocal \textbf{d}istribution information\\
		$gd(\tilde{w})$ : Global MADE model containing \textbf{g}lobal \textbf{d}istribution information\\
		$h(\bar{w}_k)$: Shallow Binary Classifier to differentiate output from $p_x$ and $q_x$ 
		$\alpha_k$: sample weights for client $k$ data\\
		\textbf{Output}:
		 $\alpha_k$: Sample weights for ${X_k}$
		\begin{algorithmic}[1]	
			\STATEx \COMMENT{Training local MADE model}		
			\FOR {$k \gets$ 1 to $K$}
			\STATEx \hskip0.5em ~\textbullet~ Client $k$: 
				\STATE Fully train $ld_k(\hat{w_k})$ on local data \{$X_k,y_k$\}.  
			\ENDFOR
			
			\STATEx 
			\STATEx \COMMENT{Training global MADE model}
			\FOR{$i \gets 1 $ to $global\_iter$}
				\FOR {$k \gets$ 1 to $K$}
					\STATEx \hskip1.5em ~\textbullet~ Client $k$:
					\STATE Update $\tilde{w}^{i-1}$ from the Aggregator
					\FOR {$j \gets 1 $ to $local\_iter$}
						\STATE  Train $gd(\tilde{w})$ on \{$X_k,y_k$\}.  
					\ENDFOR
					\STATE Sending  $gd(\tilde{w})$ to Agreegator 
				\ENDFOR
					
					\STATEx \hskip0.5em ~\textbullet~ Aggregator: 
						\STATE \hskip1.0em Aggregate   $\tilde{w}^i = \sum_{k=1}^{K} \frac{N_k}{N} \tilde{w}^i_k$
						\STATE \hskip1.0em Broatcasting $\tilde{w}^i$ to clients
			\ENDFOR
		
		\STATEx 
		\STATEx \COMMENT{Training shallow binary classifier}
		\FOR {$k \gets$ 1 to $K$}
			\STATEx \hskip1.0em Sample data from local distribution:\\ \hskip1.5em $X'^{local}_k \gets ld(X_k|\hat{w}_k)$, $y'^{local}_k \gets [0...0]$
			\STATEx \hskip1.0em Sample data from global distribution:\\ \hskip1.5em $X'^{glob}_k \gets gd(X_k|\tilde{w}_k)$ , $y'^{global}_k \gets [1...1]$
			\STATE $X'_k \gets concat(X'^{local}_k, X'^{glob}_k)$,  $y'_k \gets concat(y'^{local}_k,y'^{global}_k)$
			\STATEx \hskip0.5em ~\textbullet~ Client $k$: 
			\STATE Fully train $h(\bar{w}_k)$ on local data \{$X'_k,y'_k$\}.  
			\STATEx
			\STATEx \COMMENT{Estimate sample weight}
			\STATE $\alpha_k \gets  h(X'_k|\bar{w})[:1]$
		\ENDFOR
		
		\end{algorithmic}
	\end{flushleft}
	\label{alg:FedDisk}
\end{algorithm}

}

\Copy{sec:ablation}{
\section{Ablation Study: FedDiskAb}
\label{sec:ablation}
In this section, we examine the idea of using sample weight for non-IID data and the FedDisk sample weight effectiveness by looking at the case when the weights are derived directly from the raw data. Specifically, instead of learning sample weights from the local and global MADE model output, the weights are learned directly from the raw local and global data. To obtain the global data, we combine all client's data and randomly sample the same number of the client dataset size to avoid data imbalance. This setting variant of FedDisk (namely FedDiskAb) is an ideal case for a sample weight-based approach as it assumes to have access to the raw data. To obtain sample weights, FedDiskAb only needs to train the binary classifier on the combination of local data and global data, aiming to discriminate the two datasets. The classifier's output is used to derive the sample weight, similar to FedDisk. 

Several experiments have been conducted for FedDiskAb and other methods in Section \ref{sec:experiments}. The outcome demonstrates that both FedDiskAb and FedDisk surpass the performance of all alternative methods. This confirms the effectiveness of the sample weight-based strategy, whether acquired through learning the distribution from MADE models or directly from the data, in enhancing federated learning when faced with non-IID challenges. Furthermore, the performance of FedDisk closely aligns with that of FedDiskAb. This shows the fact that the local and global MADE models contribute significantly to the framework's ability to capture essential distribution information, much akin to the process of direct learning from the raw data. The details of the experimental results will be shown in Chapter \ref{sec:experiments}. 
}

\section{Experiments}
\label{sec:experiments}
In this section, we conduct several experiments to evaluate the proposed method on non-IID FL scenarios with three real image datasets (MNIST, Chest-Xray and FEMNIST). Our FL system goal is to learn a global classifier leveraging data from all clients. The classification accuracy is used as a metric to evaluate the performance of the proposed method. The communication cost is evaluated by counting the number of iterations needed for clients to exchange model parameters with the aggregator. We compare our method with other state-of-the-art methods, e.i., FedAvg, FedProx, FedBN, FedROD and FedPCL. 

\subsection{Datasets \& non-IID setting.}
In this Section, we describe how datasets are used in our experiments. We categorize our dataset into two groups, simulated non-IID dataset (MNIST) and real non-IID datasets (Femnist \& Chest-Xray). The first one contains images that have already been combined together so that our partitioning process is considered for sampling from the same contribution. Thus, we added different levels of noise to each client to simulate the feature skewness as inspired by settings in \cite{abs-2102-02079}, and \cite{fedpcl}. The second group's data are collected from different sources so that they are considered to be non-IID by nature. All the data are normalized and clipped to the range of [0,1] before training. Each client's data is split to 85\% and 15\% for training and testing sets, respectively. The detail of the datasets is described as follows.  

\subsubsection{Simulated non-IID: MNIST dataset}
MNIST dataset \cite{deng2012mnist} contains 60,000 (1x28x28) gray scale images of 10 digits (0-9). The number of unique output labels is 10 representing 10 digits. To mimic feature skewness, we split data equally into \parties{} partitions and add different level of noise to each client's data as inspried by the skewness simulation in \cite{abs-2102-02079}. The noise is drawn from Gaussian distribution with a mean of 0 and different values of standard deviations. More specifically, the $k^{th}$ client $(k \in [0,99])$ is added noise with the variance of $k*x/100$ where $x$ is the added noise variance.  

\subsubsection{Real non-IID: Femnist dataset}
FEMNIST dataset is downloaded from https://leaf.cmu.edu/, which is considered a benchmark dataset for real non-IID data. It contains handwritten images of 62 digits and characters (corresponding to 62 unique labels) from different writers and strokes. In this study, we randomly select \parties{} different writers (each of them owns more than 300 images to avoid overfitting) and assign their data to \parties{} clients. The average sample size of clients is 387.47, and the standard deviation is 83.04. All images are resized to a (32x32) grayscale and normalized to the range of [0,1] before inputting to models.

\subsubsection{Real non-IID: Chest Xray dataset}
The Chest-Xray dataset, which contains pneumonia and normal chest Chest-Xray images, are collected from different sources (i.e., COVID-19 \cite{covid19}, Shenzhen Hospital \cite{Shenzhen}, and University of California San Diego (UCSD) \cite{Kermany2018LabeledOC}) with different image sizes, colors and potentially taken from different medical devices. Thus, we consider this dataset non-IID by nature. After partitioning the data into \parties{} clients, the mean and standard deviation of the client sample size are 325.50 and 63.74, respectively. All images are converted to grayscale and resized to (32x32). There are two unique output labels (binary classification) to predict chest Chest-Xray images are normal or abnormal.

\subsubsection{Data examples.}
\begin{figure}[h]
	\centering
	\begin{tabular}{@{}cccc@{}}
		\includegraphics[width=.21\columnwidth]{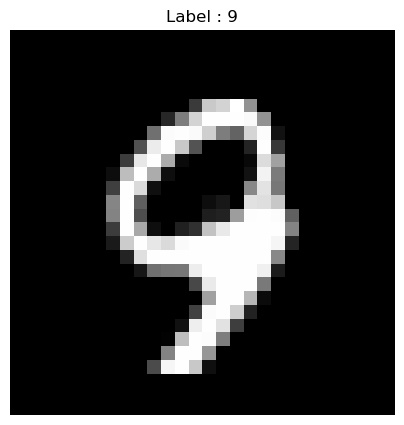} &
		\includegraphics[width=.21\columnwidth]{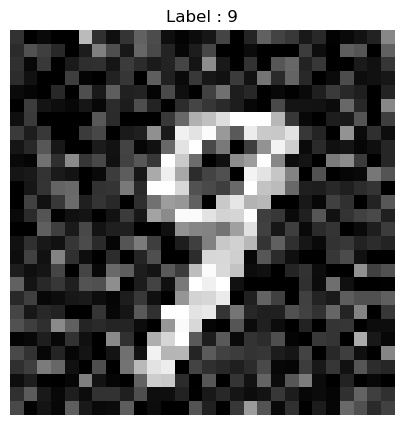} &
		\includegraphics[width=.21\columnwidth]{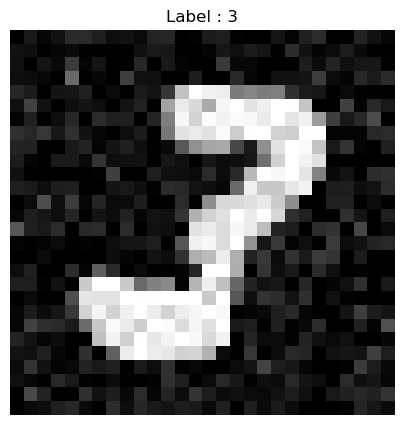} &
		\includegraphics[width=.21\columnwidth]{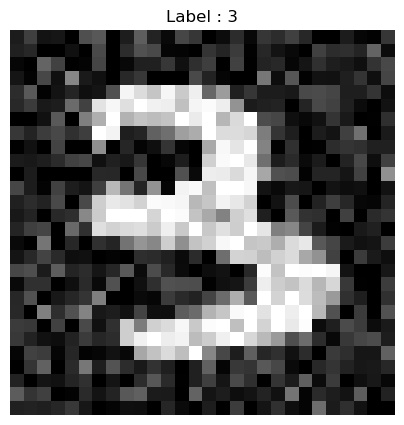} \\
		\includegraphics[width=.21\columnwidth]{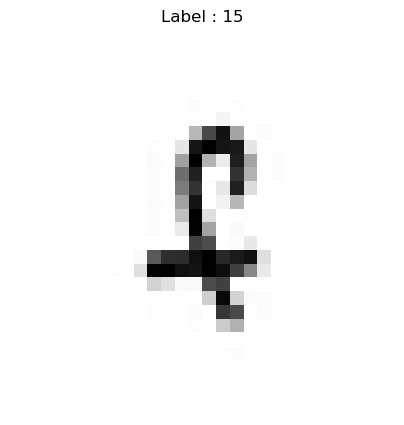} &
		\includegraphics[width=.21\columnwidth]{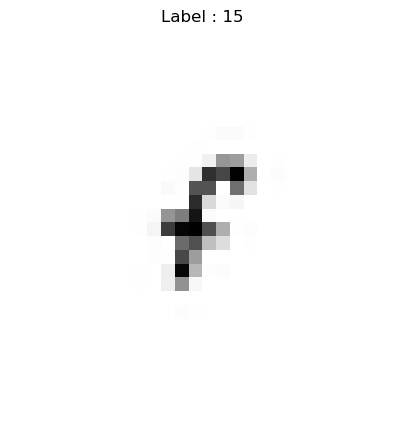} &
		\includegraphics[width=.21\columnwidth]{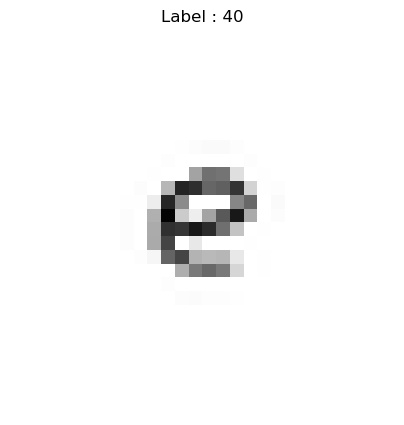} &
		\includegraphics[width=.21\columnwidth]{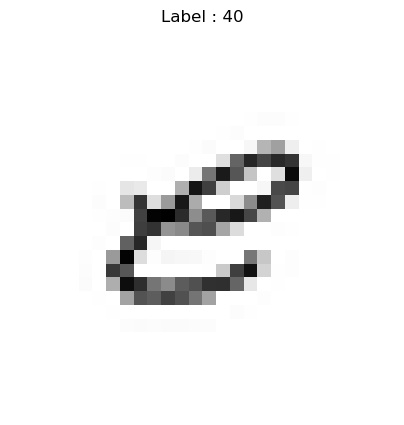} \\
		\includegraphics[width=.21\columnwidth]{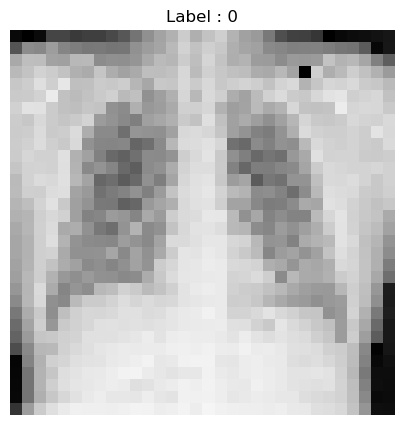} &
		\includegraphics[width=.21\columnwidth]{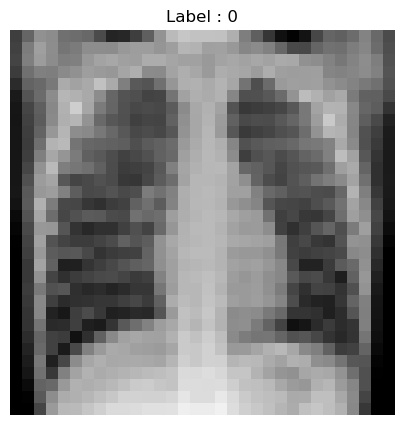} &
		\includegraphics[width=.21\columnwidth]{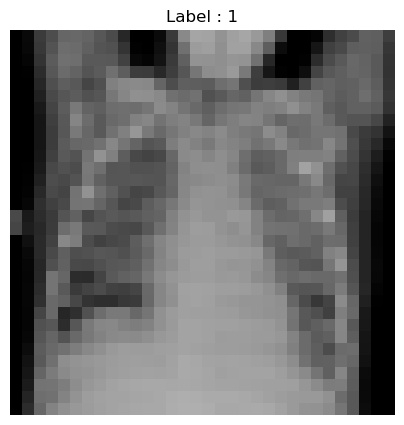} &
		\includegraphics[width=.21\columnwidth]{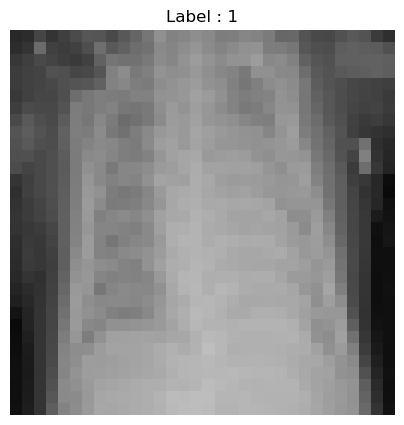} \\
	\end{tabular}
	\caption{Example images from MNIST, FEMNIST and Chest Xray datasets. They are collected from different sources and carried a veraity of resolutions, styles or conditions.}
	\label{fig:examples}
\end{figure}

Figure \ref{fig:examples} provides several sample images from the three datasets. The MNIST dataset images have various degrees of noise. Besides, the FEMNIST dataset includes images with different writing styles from various sources. The Chest-Xray dataset comprises images with varying resolutions and light conditions, etc. This makes them non-IID across clients. 

\subsection{Implementation Detail}
\subsubsection{Baselines}
We compare our methods with different methods, i.e., FedAvg, FedProx, FedBN, FedROD and FedPCL. While most implementation details are taken from the initial parameter sets in original papers, we also tune suggested parameters and report the results that give the best values. For FedROD, the results are reported for the hyper-parameter $\mu$ of 1. We also tried other values in the set {1, 5, 10, 20} and found that the results are very similar. For FedProx, we tuned the parameter $\mu$ with the candidates of {0.001, 0.01, 0.1, 1} and reported the value of 0.01, 1, 0.01 for the three datasets, Chest-Xray, FEMNIST, MNIST, respectively. 

\subsubsection{Federated Learning Classification Model} We use shallow Convolutional Neural Networks (CNN) for the classification of image datasets. The models are constructed by two 5x5 convolutional layers (32 and 32 channels for Chest-Xray, 128 and 128 channels for FEMNIST, 16 and 16 channels for MNIST). Each convolutional layer is followed by 2x2 max pooling and batch normalization layers. A fully connected layer with 16 neurons is added on the top of the models. The input and output sizes are designed to fit each dataset scenario (i.e., image size and the number of unique labels). We use stochastic gradient descent (SGD) with a learning rate of 0.01 for the optimizers. Local iterations are set to 2 for all datasets. Global iterations are set to 1500 for FEMNIST and MNIST, and 500 for Chest-Xray. 

\subsubsection{Density Estimation Model (MADE)} Density estimation models (MADE) are constructed by neural networks and the hyper-parameters are taken directly from the initial setting in the original work \cite{MADE}. Several experiments were conducted to select the optimal set of parameters which yield lowest loss value. The networks include input, output and one hidden layer. The number of neurons in the hidden layer is tuned from a value set of \{30, 50, 100, 200, 300, 400\}. The final selected number of neurons in the hidden layer are 50, 400, 30 for XRAY, FEMNIST and MNIST datasets, respectively. We noticed that using more neurons than numbers above did not significantly decrease the validation loss, thus they are the optimal settings. The model's input and output sizes are set to the flattened size of images. Specifically, the input and output size for MNIST and FEMNIST datasets is 784 with image sizes of 28x28 pixels. This number is 1024 (32x32 pixels) for Chest-Xray dataset. The maximum training iteration is set to 500, and the training process is stopped when the validation loss starts increasing. Other hyper-parameters are taken directly from \cite{MADE}.

\subsubsection{Sample Weight Approximation} In order to compute the sample weight $alpha$, we use a shallow, fully connected neural network to discriminate the density estimation output vectors coming from which of the two distribution density functions $\px{}{}$ or $\qx{}{}$. The model contains a 100-neuron hidden layer with Relu activation function. The output layer contains two neurons with Softmax activation function. All models applied a learning rate of 0.01, and SGD optimization were used in the training process. The training process is terminated if the loss function is not significantly reduced.

\subsection{Results}
\begin{figure*}[ht!]
	\centering
	\begin{subfigure}[t]{0.3\linewidth}	
		\includegraphics[width=\linewidth]{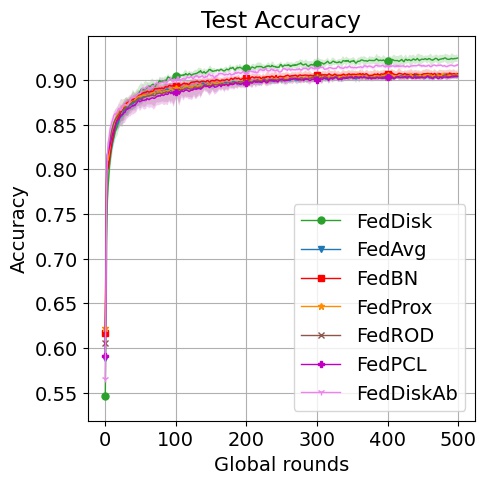}
		\caption{Chest-Xray }
		\label{fig:acc_xray}
	\end{subfigure}
	\hspace{0.01em}%
	\begin{subfigure}[t]{0.3\linewidth}
		\includegraphics[width=\linewidth]{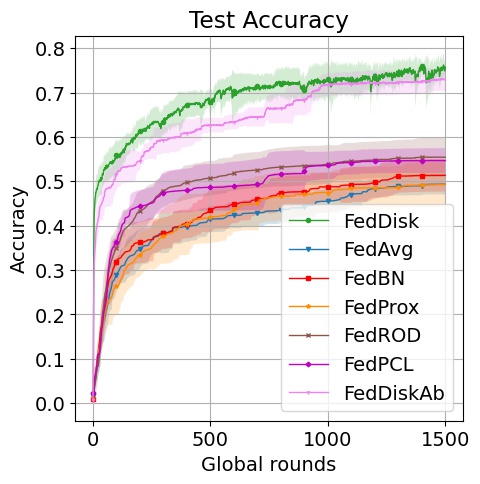}
		\caption{FEMNIST}
		\label{fig:acc_femnist}
	\end{subfigure}
	\hspace{0.01em}%
	\begin{subfigure}[t]{0.3\linewidth}	
		\includegraphics[width=\linewidth]{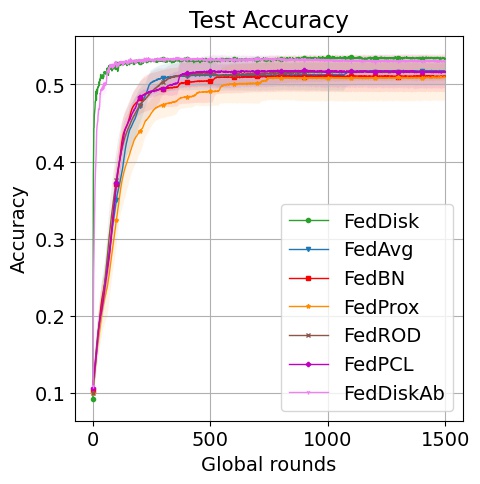}
		\caption{MNIST (NoiseVar 0.3)  }
		\label{fig:acc_mnist}
	\end{subfigure}
	\caption{Global model's average test accuracy during aggregation process. For MNIST dataset, clients' data were added noise with the mean of zero and variance of 0.3  }
	\label{fig:acc_all}
\end{figure*}

\begin{figure*}[ht!]
	\centering
	\begin{subfigure}[t]{0.3\linewidth}	
		\includegraphics[width=\linewidth]{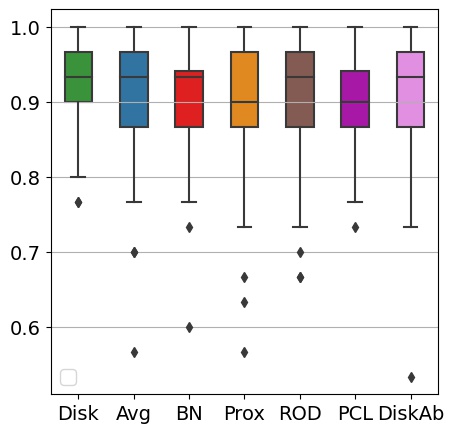}
		\caption{Chest-Xray }
		\label{fig:cl_acc_xray}
	\end{subfigure}
	\hspace{0.01em}%
	\begin{subfigure}[t]{0.3\linewidth}
		\includegraphics[width=\linewidth]{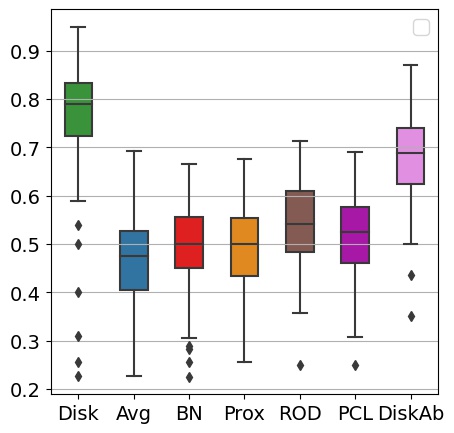}
		\caption{FEMNIST}
		\label{fig:cl_acc_femnist}
	\end{subfigure}
	\hspace{0.01em}%
	\begin{subfigure}[t]{0.3\linewidth}	
		\includegraphics[width=\linewidth]{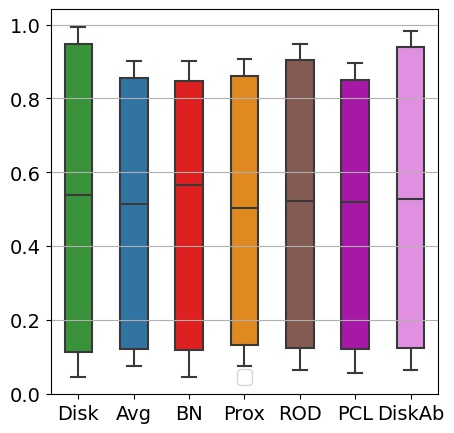}
		\caption{MNIST (NoiseVar 0.3) }
		\label{fig:cl_acc_mnist}
	\end{subfigure}
	\caption{Test accuracy percentiles, min, max and median plot of \parties{} clients for different datasets and methods. }
	\label{fig:cl_acc_all}
\end{figure*}

\subsubsection{Classification Accuracy} Figure \ref{fig:acc_all} shows the average of the \parties{} clients' testing accuracies over training iterations. The shaded regions illustrate the standard deviation over five trials. Overall, \MethodnameShort{} significantly outperforms other methods in terms of classification accuracy. For example, in Figure \ref{fig:acc_xray} for Chest-Xray dataset, \MethodnameShort{} with an accuracy of 92\% outperforms others with the highest accuracy of 90.5\%. For FEMNIST dataset (Figure \ref{fig:acc_femnist}), our method achieved an accuracy of 78\% while others only reached the maximum accuracy of 56\% (FedROD). For MNIST, \MethodnameShort{} reached the accuracy of 54.5\% while others only obtained the highest accuracy of 51.7\%. 

Figure \ref{fig:cl_acc_all} shows the descriptive statistical accuracy results of \parties{} clients on different datasets. The colored rectangles contain 50\% of client accuracies. The colored rectangular's lower and upper edges show the middle values in the first and second half of the sorted clients' accuracies (lower quartile and higher quartile). The middle dash is the median value. The upper and lower dashes represent the min and max clients' accuracies. Dots illustrate outliers. Overall, the bars for \MethodnameShort{} are higher than others, meaning that most clients archive higher accuracy. Dots are also higher (Figure \ref{fig:cl_acc_xray} and \ref{fig:cl_acc_femnist}) for \MethodnameShort{}, showing that outlier clients are also improved. Especially, the bar for FEMNIST is significantly raised for \MethodnameShort{}, indicating that the proposed method significantly improved for this dataset. It is clear that the proposed method outperforms compared methods in all experimental datasets, including real-world non-IID and simulated non-IID settings.

\begin{figure*}[ht!]
	\centering
	\begin{subfigure}[t]{0.3\linewidth}	
		\includegraphics[width=\linewidth]{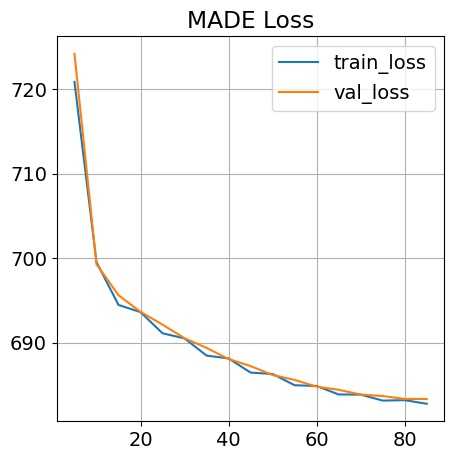}
		\caption{Chest-Xray }
		\label{fig:made_loss_xray}
	\end{subfigure}
	\hspace{0.01em}%
	\begin{subfigure}[t]{0.3\linewidth}
		\includegraphics[width=\linewidth]{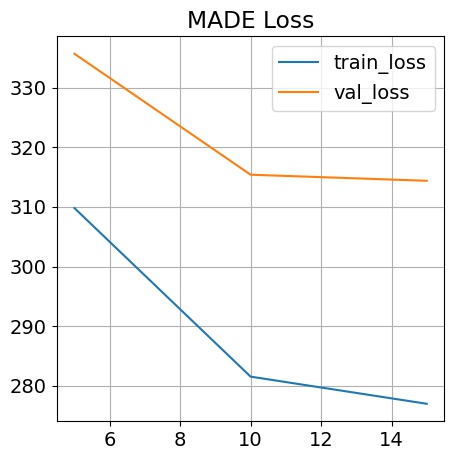}
		\caption{FEMNIST}
		\label{fig:made_loss_femnist}
	\end{subfigure}
	\hspace{0.01em}%
	\begin{subfigure}[t]{0.3\linewidth}	
		\includegraphics[width=\linewidth]{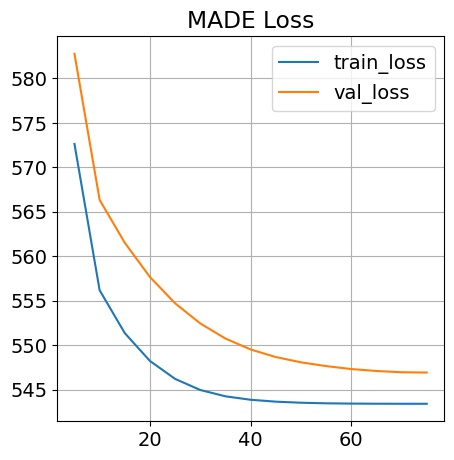}
		\caption{MNIST (NoiseVar 0.3)  }
		\label{fig:made_loss_mnist}
	\end{subfigure}
	\caption{Average validation and train losses during training the global MADE models. The training processes were stopped if the validation loss starts increasing. }
	\label{fig:made_loss}
\end{figure*}


\subsubsection{Effective Communication Rounds}
To have a fair comparision, we use ``Effective Communication Rounds" (ECR) to evaluate effective number of communication iterations for each method. On \MethodnameShort{}, ECR includes the communication rounds for exchanging MADE models and classification models. Figure \ref{fig:made_loss} show the aggregated training loss and validation loss for the global MADE model over communication rounds. The global MADE exchanging process stops when the validation loss starts increasing. For example, the proposed method needs 15 rounds for exchanging global MADE models in the case of the FEMNIST dataset (Figure \ref{fig:made_loss_femnist}). The ECR for \MethodnameShort{} in the classification phase is calculated with the number of iterations the method needs to achieve the highest value among other methods gained. Take the FEMNIST dataset experiment shown in Figure \ref{fig:cl_acc_femnist} for example, \MethodnameShort{} only needs 105 rounds to reach FedROD's accuracy at 57\% which is the highest accuracy among other experimental methods. Plus 15 rounds to exchange MADE model, \MethodnameShort{} only needs a total of 120 rounds to effectively reach the top comparison method accuracy. Since other methods don't need to exchange extra models, the ECRs are calculated by the number of rounds to exchange classification model until they reach their highest accuracy values. 

\begin{table}[!h]
        \begin{center}
                \begin{tabular}{c|c|c|c}
                        Model & Chest-Xray & FEMNIST & MNIST \\
                        \hline
                        FedDisk MADE & 80 & 15 & 70 \\
                        FedDisk Classifier & 100 & 105 & 85 \\
                        FedDisk Total & 180 & 120 & 155 \\
                        FedAvg Classifier & 450 & 1100 & 500 \\
                        FedBN Classifier & 457 & 1255 & 600 \\
                        FedProx Classifier & 440 & 1200 & 800 \\
                        FedROD Classifier & 480 & 1015 & 550 \\
                        FedPCL Classifier & 500 & 1030 & 550 \\
                \end{tabular}
        \end{center}
        \caption{The average of ``Effective Communication Rounds" for exchanging model weights between each client and the aggregator.}
        \label{tab:ecr}
\end{table}

\begin{table}[!h]
       \begin{center}
                \begin{tabular}{c|c|c|c}
                        Model & Chest-Xray & FEMNIST & MNIST \\
                        \hline
                        FedDisk MADE & 102000 & 614000 & 47000 \\
                        FedDisk Classifier & 39000 & 447000 & 12000 \\
                        FedAvg Classifier & 39000 & 447000 & 12000 \\
                        FedBN Classifier & 39000 & 447000 & 12000 \\
                        FedProx Classifier & 39000 & 447000 & 12000 \\
                        FedROD Classifier & 39000 & 447000 & 12000 \\
                        FedPCL Classifier & 39000 & 447000 & 12000 \\
                \end{tabular}
        \end{center}
        \caption{Communication overhead each round per client.}
        \label{tab:modelSize}
\end{table}

\begin{figure*}[ht!]
	\centering
	\includegraphics[width=0.7\linewidth]{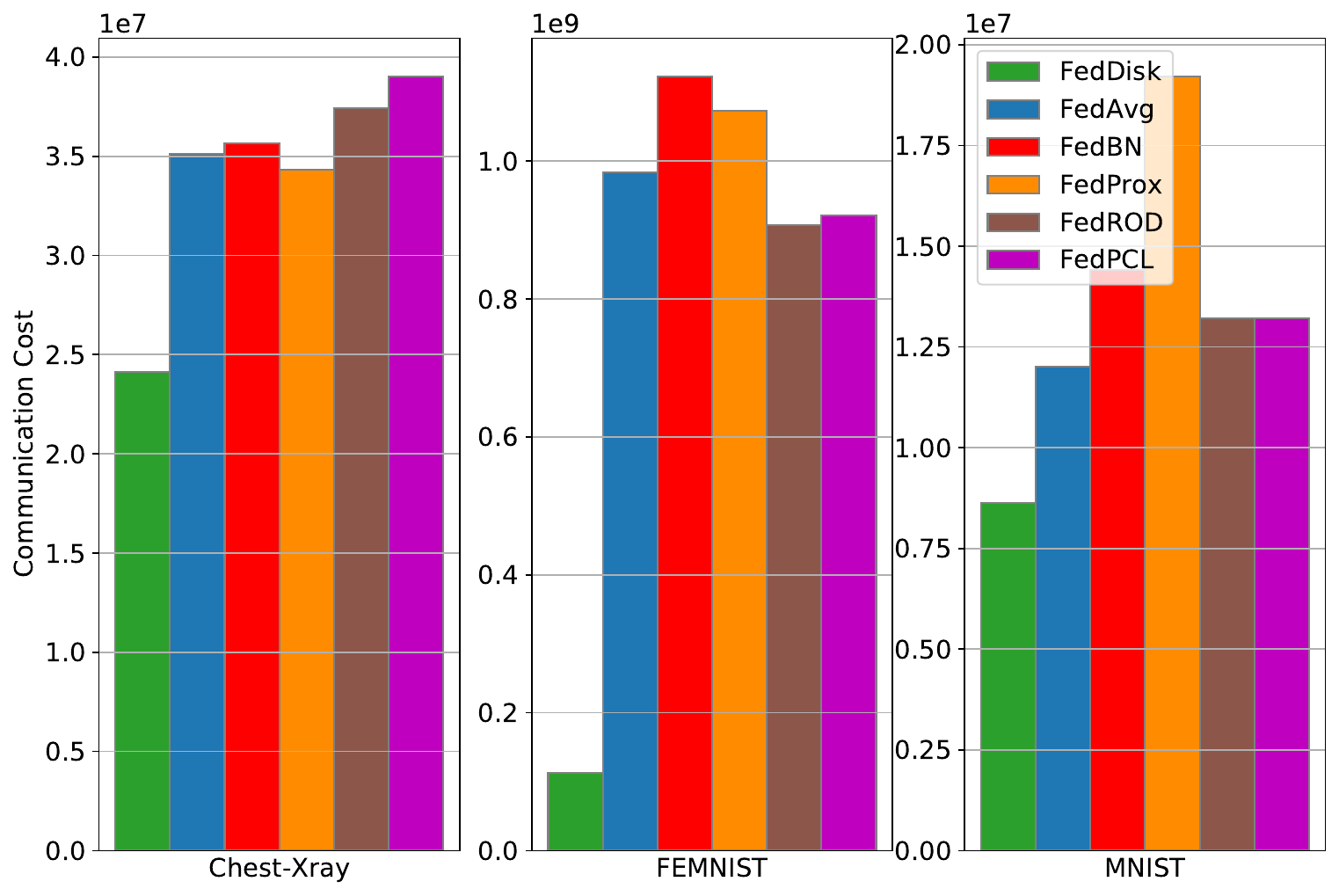}
	\caption{Summary of ``Effective Communication Cost" over 3 datasets. The Figure shows that \MethodnameShort{} is much more efficient in number of communication cost.}	
	\label{fig:communication_cost}
\end{figure*}

\begin{figure*}[h!]
	\centering
	\begin{subfigure}[]{0.3\linewidth}	
		\includegraphics[width=\linewidth]{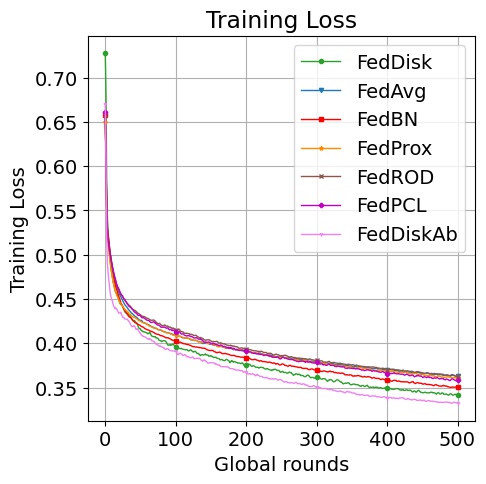}
		\caption{Chest-Xray }
	\end{subfigure}
	\hspace{0.01em}%
	\begin{subfigure}{0.3\linewidth}
		\includegraphics[width=\linewidth]{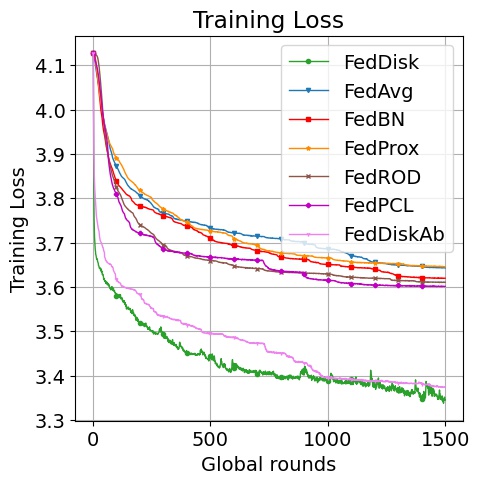}
		\caption{FEMNIST}
	\end{subfigure}
	\hspace{0.01em}%
	\begin{subfigure}{0.3\linewidth}	
		\includegraphics[width=\linewidth]{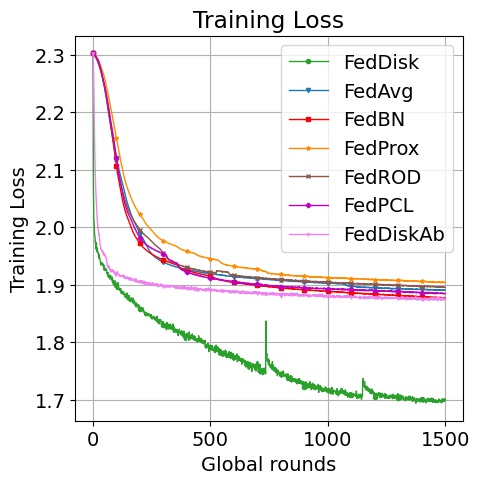}
		\caption{MNIST  }
	\end{subfigure}
	\caption{Global model's losses over communication rounds on the three datasets during training. FedDisk loss reduced much faster than other methods. }
	\label{fig:loss_all}
\end{figure*}

Table \ref{tab:ecr} shows the summary of ``Effective Communication Rounds" for the three experimental datasets. \MethodnameShort{} mechanism has two phases; one is to transfer MADE models, and the other is to exchange classifiers. The overall \MethodnameShort{} ECR comprises the communication rounds in the two phases. As shown in the Table, \MethodnameShort{} is the most effective method as it needs many fewer communication rounds to reach the highest accuracy among other methods. This is because the proposed method only needs a few number communication rounds for the global MADE model to be converged. Besides, the weight-based adjustment converges the global classification model much faster than others. For example, the \MethodnameShort{} ECR for FEMNIST is only 120 (15 for MADE model exchange plus 105 for classification model exchange), whereas others take more than 1000 iterations.

\subsubsection{Effective Communication Cost}

To have a comprehensive comparison, the communication cost is estimated for each method. The cost is comprised of communication rounds and overhead, where the overhead is measured by the size of transferred data each round between a client and the aggregator. Since the transferred data mainly contains model weights, our study uses the number of model weights to estimate the overhead size. Table \ref{tab:modelSize} shows the number of model weights in different scenarios. Note that a client must consume two costs each communication round; one is for transferring its current model weight, and another is for receiving the updated weight from the aggregator. Generally, the "effective communication cost" is then computed as follows. 
\begin{align}
    C = 2*S_{cls}*ECR_{cls} ,
\end{align}
where C is the overall communication cost for one client, $S_{cls}$ is the overhead (number of transferred classification model weights), and $ECR_{cls}$ is the number of effective communication rounds. For FedDisk, as it needs to exchange extra MADE in the first phase, the computation is adjusted as follows.
\begin{align}
    C_{FedDisk} = 2* (S_{MADE}*ECR_{MADE} + S_{cls}*ECR_{cls}) ,
\end{align}
where $S_{MADE}$ and $S_{cls}$ are model sizes needed to be transferred in phase 1 (MADE models) and phase 2 (classification models), respectively. $ECR_{MADE}$ and $ECR_{cls}$ are the corresponding effective communication rounds in the two phases. Note that for FedDisk, $ECR_{MADE} + ECR_{cls} = ECR$, and $ECR_{cls} = ECR$ for other methods.  

Figure \ref{fig:communication_cost} summarizes effective communication cost. Noticeably, the communication cost reduced significantly for the FEMNIST dataset under the proposed method, 8 times, from the second lowest cost method of 907,000,000 (FedROD, brown column) to 112,000,000 (FedDisk, green column). The communication cost reduction trend is also applied to other experimental datasets, Chest-Xray (1.4 times) and MNIST (1.6 times). Thus, the proposed method improves accuracy and dramatically reduces communication costs, one of the most critical concerns in Federated Learning. This is because the loss function were reduced faster as we adjusted using the sample weights. Figure \ref{fig:loss_all} demonstrates the global loss values during training. For FedDisk, the sample weights effectively affect the optimization function, and the loss reduces much faster, and the accuracies proportionally increase faster.   

\Copy{sec:discussion}{
\subsection{Discusion}
The proposed method offers a holistic improvement over existing federated learning methods. Its combination of enhanced accuracy and reduced communication costs signifies its effectiveness across a variety of datasets and scenarios. For example, the accuray can be increased by 22\% and the communication time reduced by 8 times for FEMNIST dataset. 
While FedDisk exhibits remarkable performance across various metrics in federated learning scenarios, it's important to acknowledge a drawback associated with its larger model size compared to some other methods as described in Table \ref{tab:modelSize}. The increased model size leads to higher space occupation on client devices, which can have implications for light weight devices with limited storage capacity. Hence, the suggested approach could be well-suited for the cross-silo scenario, typically characterized by clients having ample data and adequate computational capabilities. 
Overal, the proposed approach provides an avenue to address critical challenges in federated learning, making it a promising option for real-world applications. 
}

\section{Privacy leakage analysis}
\label{sec:privacyAnalysis}
In this section, we discuss the privacy leakage of our method compared to the conventional FL. Similar to many other works, we utilize additional information to alleviate the negative impact of non-IID data, i.e., parameters of MADE models. However, these parameters might contain distribution information of clients’ data. However, we prove that the more clients are involved in the FL training process, the less our extra information is leaked. The detail is described as follows. 

Assume each client samples their own data point ${\hat Z}_k \sim Q_k$ independently, and let  $\Theta$ be a random variable taking values in $\llbracket 1,K \rrbracket$ with $\prob[\Theta = k] = \kappa_k$ and independent of ${\hat Z}_k$ for each $k \in \llbracket 1,K \rrbracket$. Note that ${\hat Z}_\Theta \sim P$, and one may quantify the privacy leakage of client $k$'s data through the knowledge of $P$ by the mutual  information between ${\hat Z}_k$ and ${\hat Z}_\Theta$, as given by
\begin{equation}
	\begin{aligned}
		I({\hat Z}_k;{\hat Z}_\Theta) & \leq I({\hat Z}_k;{\hat Z}_\Theta, \Theta)
		=  I({\hat Z}_k; \Theta) + I({\hat Z}_k;{\hat Z}_\Theta | \Theta)\\
		&= I({\hat Z}_k;{\hat Z}_\Theta | \Theta) = \sum_{i=1}^K \prob[\Theta = i] I({\hat Z}_k;{\hat Z}_i) \\
		&= \kappa_k H({\hat Z}_k).
	\end{aligned}
\end{equation}
In other words, the privacy leakage is proportional to $\kappa_k$, which decreases to $0$ as long as $\kappa_k = O(1/K)$ and $K \rightarrow \infty$.

\section{Conclusion}
\label{sec:conclusion}
In this work, we have proposed an FL method to tackle the issue of data distribution skewness. The technique utilizes an FL framework and a neural network-based density estimation model to derive training sample weights. This helps to adjust the individual distribution without revealing clients' raw data. Thus, the global model loss is converged faster and more accurately. The experimental results show that the proposed method improves FL accuracy and significantly reduces communication costs. We also provide a privacy analysis for the extra information used in \MethodnameShort{} (i.e., the parameters of MADE models) and prove that the leakage information becomes less important when the number of clients increases. To advance our research, we intend to fine-tune the parameters of the MADE model and investigate various distribution models in order to potentially achieve improved outcomes and have a greater impact on real-world applications. More analysis on privacy vulnerabilities and potential attacks will also be our future research.

\bibliographystyle{IEEEtran}
\bibliography{egbib}

\begin{thebibliography}{10}
\providecommand{\url}[1]{#1}
\csname url@samestyle\endcsname
\providecommand{\newblock}{\relax}
\providecommand{\bibinfo}[2]{#2}
\providecommand{\BIBentrySTDinterwordspacing}{\spaceskip=0pt\relax}
\providecommand{\BIBentryALTinterwordstretchfactor}{4}
\providecommand{\BIBentryALTinterwordspacing}{\spaceskip=\fontdimen2\font plus
\BIBentryALTinterwordstretchfactor\fontdimen3\font minus
  \fontdimen4\font\relax}
\providecommand{\BIBforeignlanguage}[2]{{%
\expandafter\ifx\csname l@#1\endcsname\relax
\typeout{** WARNING: IEEEtran.bst: No hyphenation pattern has been}%
\typeout{** loaded for the language `#1'. Using the pattern for}%
\typeout{** the default language instead.}%
\else
\language=\csname l@#1\endcsname
\fi
#2}}
\providecommand{\BIBdecl}{\relax}
\BIBdecl

\bibitem{OriginFL}
\BIBentryALTinterwordspacing
B.~McMahan, E.~Moore, D.~Ramage, S.~Hampson, and B.~A.~y. Arcas,
  ``{Communication-Efficient Learning of Deep Networks from Decentralized
  Data},'' in \emph{Proceedings of the 20th International Conference on
  Artificial Intelligence and Statistics}, ser. Proceedings of Machine Learning
  Research, A.~Singh and J.~Zhu, Eds., vol.~54.\hskip 1em plus 0.5em minus
  0.4em\relax PMLR, 20--22 Apr 2017, pp. 1273--1282. [Online]. Available:
  \url{https://proceedings.mlr.press/v54/mcmahan17a.html}
\BIBentrySTDinterwordspacing

\bibitem{Zhao2018FederatedLW}
Y.~Zhao, M.~Li, L.~Lai, N.~Suda, D.~Civin, and V.~Chandra, ``Federated learning
  with non-iid data,'' \emph{ArXiv}, vol. abs/1806.00582, 2018.

\bibitem{cifar10}
\BIBentryALTinterwordspacing
A.~Krizhevsky, V.~Nair, and G.~Hinton, ``Cifar-10 (canadian institute for
  advanced research).'' [Online]. Available:
  \url{http://www.cs.toronto.edu/~kriz/cifar.html}
\BIBentrySTDinterwordspacing

\bibitem{kws}
P.~Warden, ``Speech commands: A dataset for limited-vocabulary speech
  recognition,'' 2018.

\bibitem{abs-1811-03604}
\BIBentryALTinterwordspacing
A.~Hard, K.~Rao, R.~Mathews, F.~Beaufays, S.~Augenstein, H.~Eichner, C.~Kiddon,
  and D.~Ramage, ``Federated learning for mobile keyboard prediction,''
  \emph{CoRR}, vol. abs/1811.03604, 2018. [Online]. Available:
  \url{http://arxiv.org/abs/1811.03604}
\BIBentrySTDinterwordspacing

\bibitem{yadav_federated_2022}
S.~P. Yadav, B.~S. Bhati, D.~P. Mahato, S.~Kumar, and {SpringerLink (Online
  service)}, \emph{\BIBforeignlanguage{English}{Federated {Learning} for {IoT}
  {Applications}}}.\hskip 1em plus 0.5em minus 0.4em\relax Cham: Springer
  International Publishing Imprint Springer., 2022, oCLC: 1295622946.

\bibitem{feki_federated_2021}
\BIBentryALTinterwordspacing
I.~Feki, S.~Ammar, Y.~Kessentini, and K.~Muhammad,
  ``\BIBforeignlanguage{en}{Federated learning for {COVID}-19 screening from
  {Chest} {X}-ray images},'' \emph{\BIBforeignlanguage{en}{Applied Soft
  Computing}}, vol. 106, p. 107330, Jul. 2021. [Online]. Available:
  \url{https://linkinghub.elsevier.com/retrieve/pii/S1568494621002532}
\BIBentrySTDinterwordspacing

\bibitem{zhai_dynamic_2021}
\BIBentryALTinterwordspacing
S.~Zhai, X.~Jin, L.~Wei, H.~Luo, and M.~Cao, ``Dynamic {Federated} {Learning}
  for {GMEC} {With} {Time}-{Varying} {Wireless} {Link},'' \emph{IEEE Access},
  vol.~9, pp. 10\,400--10\,412, 2021. [Online]. Available:
  \url{https://ieeexplore.ieee.org/document/9317862/}
\BIBentrySTDinterwordspacing

\bibitem{distributedQuantum}
A.~K. Singh, D.~Saxena, J.~Kumar, and V.~Gupta, ``A quantum approach towards
  the adaptive prediction of cloud workloads,'' \emph{IEEE Transactions on
  Parallel and Distributed Systems}, vol.~32, no.~12, pp. 2893--2905, 2021.

\bibitem{hierarchicalFL}
W.~Y.~B. Lim, J.~S. Ng, Z.~Xiong, J.~Jin, Y.~Zhang, D.~Niyato, C.~Leung, and
  C.~Miao, ``Decentralized edge intelligence: A dynamic resource allocation
  framework for hierarchical federated learning,'' \emph{IEEE Transactions on
  Parallel and Distributed Systems}, vol.~33, no.~3, pp. 536--550, 2022.

\bibitem{ZHU2021371}
\BIBentryALTinterwordspacing
H.~Zhu, J.~Xu, S.~Liu, and Y.~Jin, ``Federated learning on non-iid data: A
  survey,'' \emph{Neurocomputing}, vol. 465, pp. 371--390, 2021. [Online].
  Available:
  \url{https://www.sciencedirect.com/science/article/pii/S0925231221013254}
\BIBentrySTDinterwordspacing

\bibitem{Sahu2018OnTC}
A.~K. Sahu, T.~Li, M.~Sanjabi, M.~Zaheer, A.~S. Talwalkar, and V.~Smith, ``On
  the convergence of federated optimization in heterogeneous networks,''
  \emph{ArXiv}, vol. abs/1812.06127, 2018.

\bibitem{9392310}
S.~Itahara, T.~Nishio, Y.~Koda, M.~Morikura, and K.~Yamamoto,
  ``Distillation-based semi-supervised federated learning for
  communication-efficient collaborative training with non-iid private data,''
  \emph{IEEE Transactions on Mobile Computing}, no.~01, pp. 1--1, mar 2021.

\bibitem{abs_1905_06641}
L.~Liu, J.~Zhang, S.~Song, and K.~B. Letaief, ``Edge-assisted hierarchical
  federated learning with non-iid data,'' \emph{CoRR}, vol. abs/1905.06641,
  2019.

\bibitem{Shen2020FederatedML}
T.~Shen, J.~Zhang, X.~Jia, F.~Zhang, G.~Huang, P.~Zhou, F.~Wu, and C.~Wu,
  ``Federated mutual learning,'' \emph{ArXiv}, vol. abs/2006.16765, 2020.

\bibitem{9155494}
H.~Wang, Z.~Kaplan, D.~Niu, and B.~Li, ``Optimizing federated learning on
  non-iid data with reinforcement learning,'' in \emph{IEEE INFOCOM 2020 - IEEE
  Conference on Computer Communications}, 2020, pp. 1698--1707.

\bibitem{abs-2102-02079}
\BIBentryALTinterwordspacing
Q.~Li, Y.~Diao, Q.~Chen, and B.~He, ``Federated learning on non-iid data silos:
  An experimental study,'' \emph{CoRR}, vol. abs/2102.02079, 2021. [Online].
  Available: \url{https://arxiv.org/abs/2102.02079}
\BIBentrySTDinterwordspacing

\bibitem{abs-2005-11418}
\BIBentryALTinterwordspacing
X.~Zhang, M.~Hong, S.~V. Dhople, W.~Yin, and Y.~Liu, ``Fedpd: {A} federated
  learning framework with optimal rates and adaptivity to non-iid data,''
  \emph{CoRR}, vol. abs/2005.11418, 2020. [Online]. Available:
  \url{https://arxiv.org/abs/2005.11418}
\BIBentrySTDinterwordspacing

\bibitem{nofearofheterogeneity}
M.~Luo, F.~Chen, D.~Hu, Y.~Zhang, J.~Liang, and J.~Feng, ``No fear of
  heterogeneity: Classifier calibration for federated learning with non-iid
  data,'' in \emph{Advances in Neural Information Processing Systems},
  A.~Beygelzimer, Y.~Dauphin, P.~Liang, and J.~W. Vaughan, Eds., 2021.

\bibitem{distillationFL}
Z.~Zhu, J.~Hong, and J.~Zhou, ``Data-free knowledge distillation for
  heterogeneous federated learning,'' 2021.

\bibitem{li2021fedbn}
\BIBentryALTinterwordspacing
X.~Li, M.~JIANG, X.~Zhang, M.~Kamp, and Q.~Dou, ``Fed{BN}: Federated learning
  on non-{IID} features via local batch normalization,'' in \emph{International
  Conference on Learning Representations}, 2021. [Online]. Available:
  \url{https://openreview.net/forum?id=6YEQUn0QICG}
\BIBentrySTDinterwordspacing

\bibitem{FedProx}
\BIBentryALTinterwordspacing
A.~K. Sahu, T.~Li, M.~Sanjabi, M.~Zaheer, A.~Talwalkar, and V.~Smith, ``On the
  convergence of federated optimization in heterogeneous networks,''
  \emph{CoRR}, vol. abs/1812.06127, 2018. [Online]. Available:
  \url{http://arxiv.org/abs/1812.06127}
\BIBentrySTDinterwordspacing

\bibitem{fednova}
J.~Wang, Q.~Liu, H.~Liang, G.~Joshi, and H.~Vincent~Poor,
  ``\BIBforeignlanguage{"English (US)"}{"tackling the objective inconsistency
  problem in heterogeneous federated optimization"},''
  \emph{\BIBforeignlanguage{"English (US)"}{"Advances in Neural Information
  Processing Systems"}}, vol. "2020-December", "2020".

\bibitem{fedma}
\BIBentryALTinterwordspacing
H.~Wang, M.~Yurochkin, Y.~Sun, D.~Papailiopoulos, and Y.~Khazaeni, ``Federated
  learning with matched averaging,'' in \emph{International Conference on
  Learning Representations}, 2020. [Online]. Available:
  \url{https://openreview.net/forum?id=BkluqlSFDS}
\BIBentrySTDinterwordspacing

\bibitem{AFL}
\BIBentryALTinterwordspacing
M.~Mohri, G.~Sivek, and A.~T. Suresh, ``Agnostic federated learning,'' in
  \emph{Proceedings of the 36th International Conference on Machine Learning},
  ser. Proceedings of Machine Learning Research, K.~Chaudhuri and
  R.~Salakhutdinov, Eds., vol.~97.\hskip 1em plus 0.5em minus 0.4em\relax PMLR,
  09--15 Jun 2019, pp. 4615--4625. [Online]. Available:
  \url{https://proceedings.mlr.press/v97/mohri19a.html}
\BIBentrySTDinterwordspacing

\bibitem{pfnm}
\BIBentryALTinterwordspacing
M.~Yurochkin, M.~Agarwal, S.~Ghosh, K.~Greenewald, N.~Hoang, and Y.~Khazaeni,
  ``{B}ayesian nonparametric federated learning of neural networks,'' in
  \emph{Proceedings of the 36th International Conference on Machine Learning},
  ser. Proceedings of Machine Learning Research, K.~Chaudhuri and
  R.~Salakhutdinov, Eds., vol.~97.\hskip 1em plus 0.5em minus 0.4em\relax PMLR,
  09--15 Jun 2019, pp. 7252--7261. [Online]. Available:
  \url{https://proceedings.mlr.press/v97/yurochkin19a.html}
\BIBentrySTDinterwordspacing

\bibitem{FedRod}
\BIBentryALTinterwordspacing
H.-Y. Chen and W.-L. Chao, ``On bridging generic and personalized federated
  learning for image classification,'' in \emph{International Conference on
  Learning Representations}, 2022. [Online]. Available:
  \url{https://openreview.net/forum?id=I1hQbx10Kxn}
\BIBentrySTDinterwordspacing

\bibitem{fedpcl}
Y.~Tan, G.~Long, J.~Ma, L.~Liu, T.~Zhou, and J.~Jiang, ``Federated learning
  from pre-trained models: A contrastive learning approach,'' 2022.

\bibitem{fedcl}
M.~Wang, J.~Guo, and W.~Jia, ``Fedcl: Federated multi-phase curriculum learning
  to synchronously correlate user heterogeneity,'' 2023.

\bibitem{fedDNA}
J.-H. Duan, W.~Li, and S.~Lu, \emph{FedDNA: Federated Learning with Decoupled
  Normalization-Layer Aggregation for Non-IID Data}, 09 2021, pp. 722--737.

\bibitem{survey}
\BIBentryALTinterwordspacing
T.~Li, A.~K. Sahu, M.~Zaheer, M.~Sanjabi, A.~Talwalkar, and V.~Smith,
  ``Federated optimization in heterogeneous networks,'' in \emph{MLSys}, 2020.
  [Online]. Available: \url{https://proceedings.mlsys.org/book/316.pdf}
\BIBentrySTDinterwordspacing

\bibitem{MADE}
\BIBentryALTinterwordspacing
M.~Germain, K.~Gregor, I.~Murray, and H.~Larochelle, ``Made: Masked autoencoder
  for distribution estimation,'' in \emph{Proceedings of the 32nd International
  Conference on Machine Learning}, ser. Proceedings of Machine Learning
  Research, F.~Bach and D.~Blei, Eds., vol.~37.\hskip 1em plus 0.5em minus
  0.4em\relax Lille, France: PMLR, 07--09 Jul 2015, pp. 881--889. [Online].
  Available: \url{https://proceedings.mlr.press/v37/germain15.html}
\BIBentrySTDinterwordspacing

\bibitem{noauthor_empirical_2021}
C.~Jin, L.~T. Liu, R.~Ge, and M.~I. Jordan, ``On the local minima of the
  empirical risk,'' in \emph{Advances in Neural Information Processing
  Systems}, S.~Bengio, H.~Wallach, H.~Larochelle, K.~Grauman, N.~Cesa-Bianchi,
  and R.~Garnett, Eds., vol.~31.\hskip 1em plus 0.5em minus 0.4em\relax Curran
  Associates, Inc., 2018.

\bibitem{densityratio}
\BIBentryALTinterwordspacing
A.~Menon and C.~S. Ong, ``Linking losses for density ratio and
  class-probability estimation,'' in \emph{Proceedings of The 33rd
  International Conference on Machine Learning}, ser. Proceedings of Machine
  Learning Research, M.~F. Balcan and K.~Q. Weinberger, Eds., vol.~48.\hskip
  1em plus 0.5em minus 0.4em\relax New York, New York, USA: PMLR, 20--22 Jun
  2016, pp. 304--313. [Online]. Available:
  \url{https://proceedings.mlr.press/v48/menon16.html}
\BIBentrySTDinterwordspacing

\bibitem{deng2012mnist}
L.~Deng, ``The mnist database of handwritten digit images for machine learning
  research,'' \emph{IEEE Signal Processing Magazine}, vol.~29, no.~6, pp.
  141--142, 2012.

\bibitem{covid19}
\BIBentryALTinterwordspacing
M.~E.~H. Chowdhury, T.~Rahman, A.~Khandakar, R.~Mazhar, M.~A. Kadir, Z.~B.
  Mahbub, K.~R. Islam, M.~S. Khan, A.~Iqbal, N.~Al{-}Emadi, and M.~B.~I. Reaz,
  ``Can {AI} help in screening viral and {COVID-19} pneumonia?'' \emph{CoRR},
  vol. abs/2003.13145, 2020. [Online]. Available:
  \url{https://arxiv.org/abs/2003.13145}
\BIBentrySTDinterwordspacing

\bibitem{Shenzhen}
S.~Jaeger, S.~Candemir, S.~Antani, Y.-X. Wáng, P.-X. Lu, and G.~Thoma, ``Two
  public chest x-ray datasets for computer-aided screening of pulmonary
  diseases,'' \emph{Quantitative imaging in medicine and surgery}, vol.~4, pp.
  475--7, 12 2014.

\bibitem{Kermany2018LabeledOC}
D.~S. Kermany, K.~Zhang, and M.~H. Goldbaum, ``Labeled optical coherence
  tomography (oct) and chest x-ray images for classification,'' 2018.

\bibitem{FLviaIntel}
Z.~Wang, J.~Qiu, Y.~Zhou, Y.~Shi, L.~Fu, W.~Chen, and K.~B. Letaief,
  ``Federated learning via intelligent reflecting surface,'' \emph{IEEE
  Transactions on Wireless Communications}, vol.~21, no.~2, pp. 808--822, 2022.

\end{thebibliography}

\begin{IEEEbiography}[{\includegraphics[width=0.9\linewidth,clip,keepaspectratio]{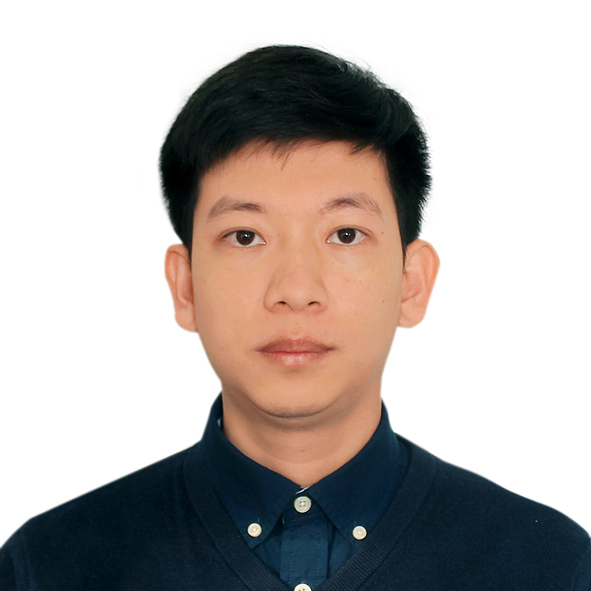}}]{Hung Nguyen}
	received his Ph.D. degree in Department of Electrical Engineering, University of South Florida, FL, USA, in 2023. His current research interests include machine learning, artificial intelligence, federated learning, cyber security, privacy enhancing technologies. Hung is a member of IEEE.
\end{IEEEbiography}

\begin{IEEEbiography}[{\includegraphics[width=0.9\linewidth,clip,keepaspectratio]{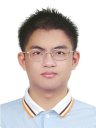}}]{Pei-Yuan Wu}
	Pei-Yuan Wu (Member, IEEE) was born in Taipei,
	Taiwan, in 1987. He received the B.S.E. degree
	in electrical engineering from National Taiwan
	University, Taipei, in 2009, and the M.A. and Ph.D.
	degrees in electrical engineering from Princeton
	University, Princeton, NJ, USA, in 2012 and 2015,
	respectively.
\end{IEEEbiography}

\begin{IEEEbiography}[{\includegraphics[width=0.9\linewidth,clip,keepaspectratio]{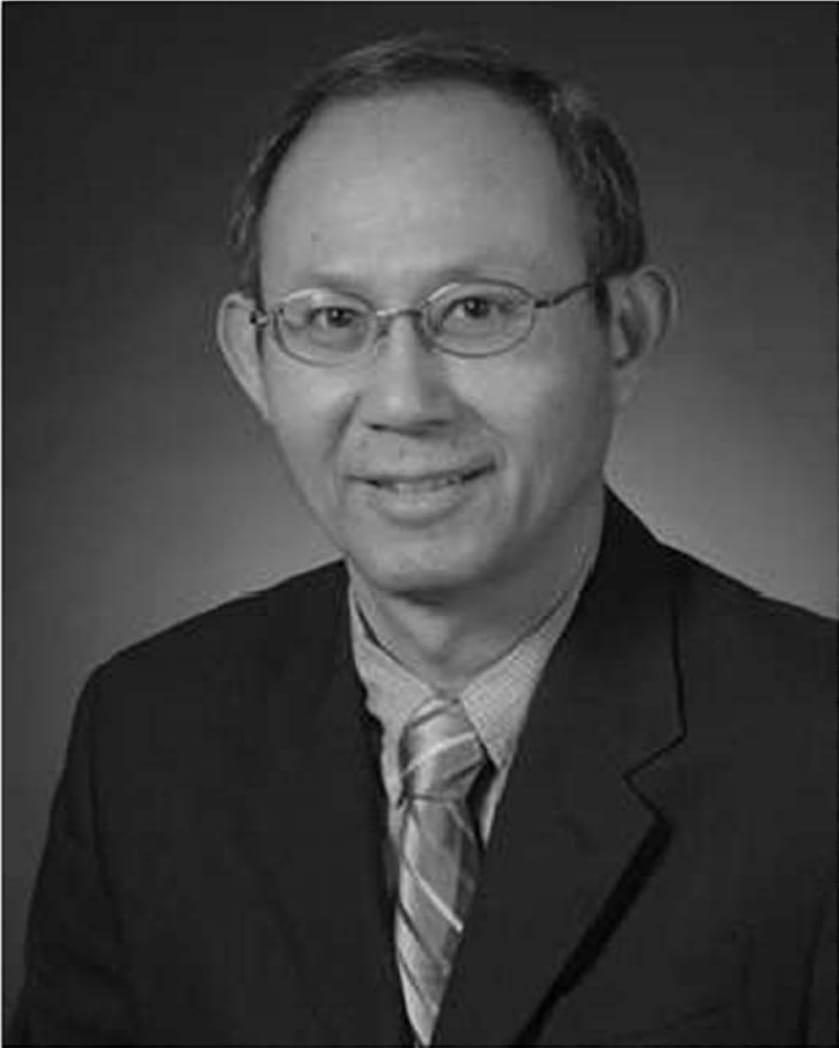}}]{J. Morris Chang}
	(SM'08) is a professor in the Department of Electrical Engineering at the University of South Florida. He received his Ph.D. degree from the North Carolina State University. His past industrial experiences include positions at Texas Instruments, Microelectronic Center of North Carolina and AT\&T Bell Labs. He received the University Excellence in Teaching Award at Illinois Institute of Technology in 1999. His research interests include: cyber-security, wireless networks, and energy efficient computer systems. In the last six years, his research projects on cyber-security have been funded by DARPA. Currently, he is leading a DARPA project under Brandeis program focusing on privacy-preserving computation over Internet. He is a handling editor of Journal of Microprocessors and Microsystems and an editor of IEEE IT Professional. He is a senior member of IEEE.
\end{IEEEbiography}


\end{document}